\documentclass[a4paper,fleqn]{cas-dc}

\usepackage{cite}
\usepackage{listings}
\usepackage{placeins}
\usepackage{amsmath,amssymb,amsfonts}
\usepackage{algorithmic}
\usepackage[linesnumbered,algoruled,boxed,lined]{algorithm2e}
\usepackage{graphicx}
\usepackage{textcomp}
\usepackage{xcolor}
\usepackage{tabularx}
\usepackage{multirow}
\usepackage{booktabs}
\usepackage{adjustbox}
\usepackage{caption}
\usepackage{subcaption}
\usepackage{lineno}
\usepackage{tcolorbox}
\usepackage{xspace}
\usepackage{hyperref}

\def\BibTeX{{\rm B\kern-.05em{\sc i\kern-.025em b}\kern-.08em
    T\kern-.1667em\lower.7ex\hbox{E}\kern-.125emX}}

\makeatletter
\g@addto@macro{\@algocf@init}{\SetKwInOut{Parameter}{Parameters}}
\makeatother

\newcommand{\tool}{\textsc{RAG-Tester}\xspace}

\newcommand{\chunk}[2]{%
\fcolorbox{black}{yellow}{\bfseries\sffamily\scriptsize#1}%
{$\blacktriangleright$#2$\blacktriangleleft$}%
}

\newcommand{\aitor}[1]{\chunk{Aitor}{\textbf{\textcolor{red}{\textsl{#1}}}}}
\newcommand{\ange}[1]{\chunk{Ange}{\textbf{\textcolor{blue}{\textsl{#1}}}}}

\begin{document}
\let\WriteBookmarks\relax
\def\floatpagepagefraction{1}
\def\textpagefraction{.001}

\title [mode = title]{\tool: Automated End-to-End Testing of Retrieval-Augmented Large Language Models}

\shorttitle{Automated End-to-End Testing of Retrieval-Augmented Large Language Models}
\shortauthors{Ange Maiztegui, Jon Ayerdi, Miren Illarramendi, Aitor Arrieta}

\author[1]{Ange Maiztegui}
\ead{ange.maiztegui@alumni.mondragon.edu}

\author[1]{Jon Ayerdi}
\ead{jayerdi@mondragon.edu}

\author[1]{Miren Illarramendi}
\ead{millarramendi@mondragon.edu}

\author[1]{Aitor Arrieta}
\cormark[1]
\ead{aarrieta@mondragon.edu}

\affiliation[1]{
    organization={Mondragon University},
    city={Arrasate-Mondragon},
    state={Gipuzkoa},
    country={Spain}
}

\cortext[cor1]{Corresponding author}

\begin{keywords}
LLM, Testing, RAG
\end{keywords}
\maketitle

\begin{abstract}
\space
\textbf{Context:}
Retrieval-Augmented Generation (RAG) enables Large Language Models (LLMs) to incorporate external, up-to-date, domain-specific, or proprietary information at inference time. However, the reliability of a RAG system depends on the interaction between multiple components, including the generative model, embedding model, retrieval mechanism, and prompt construction strategy. Different combinations of these components may therefore exhibit substantially different failure behavior, motivating systematic and automated testing.

\textbf{Objectives:}
This work aims to automate the end-to-end testing of RAG-enabled LLMs and to generate test cases capable of exposing failures across different combinations of LLMs and embedding models.

\textbf{Methods:}
We propose \tool, an automated testing approach comprising four stages: generation of retrieval documents, generation of test inputs and expected outputs, execution of the generated tests, and automated evaluation using an LLM as a judge. The test-generation strategy incorporates complex document passages, unsupported queries, and document-coverage criteria. We evaluate \tool using eight LLMs and six embedding models, resulting in 24 compatible LLM--embedding configurations, and compare it with a baseline test-input generator.

\textbf{Results:}
Across 72,000 test executions, \tool detected 21,633 failures, compared with 20,293 detected by the baseline, representing a 6.6\% increase. \tool outperformed the baseline in 20 of the 24 evaluated configurations and exposed failures including retrieval inaccuracies, unsupported answers, incomplete use of retrieved context, and difficulties interpreting complex passages.

\textbf{Conclusion:}
The results show that automated, coverage-oriented test generation can effectively expose failures arising from the interaction between retrieval and generation components in RAG systems. \tool provides an end-to-end mechanism for comparing LLM--embedding configurations and assessing their reliability before deployment.
\end{abstract}



\section{Introduction}

Recent advancements in Large Language Models (LLMs) have led to substantial improvements in natural language understanding and generation across a wide range of tasks, including question answering, code generation, conversational agents, and decision-support systems. Despite these advances, traditional LLMs primarily rely on knowledge encoded in their parameters, making them inherently limited by outdated training data, prone to hallucinations, and unable to provide verifiable, context-specific information.

Retrieval-Augmented Generation (RAG)~\cite{lewis2020rag} has emerged as a promising paradigm to mitigate these limitations by integrating external knowledge retrieval mechanisms with neural text generation. By incorporating relevant documents at inference time, RAG systems can produce responses that are more accurate, up-to-date, and grounded in source material. As a result, RAG-based LLMs are increasingly adopted in high-impact domains such as law, medicine, and enterprise knowledge management, where reliability and traceability are critical.

However, the development of RAG-enabled LLMs introduces new challenges from a software engineering perspective. Unlike standalone LLMs, the correctness of a RAG system depends on multiple interacting components, including embedding models, retrieval pipelines, ranking mechanisms, and prompt construction strategies. Consequently, failures may arise from different sources, such as inaccurate retrieval, incomplete context integration, or hallucinated responses. This complexity makes systematic and automated testing of RAG systems paramount, yet non-trivial.

Existing approaches for evaluating RAG systems largely rely on benchmark-based assessments or scenario-specific evaluation frameworks~\cite{zhu2024rageval,tang2024multihop, pipitone2024legalbench, friel2024ragbench, yang2024crag}. While benchmarking techniques provide standardized comparisons, they are inherently limited in coverage, as they focus on predefined datasets or narrow application scenarios. As a result, they often fail to expose diverse failure modes, particularly those arising from the interaction between retrieval and generation components. More importantly, current approaches lack end-to-end automated testing mechanisms capable of systematically generating test inputs, defining expected outputs, and evaluating system behavior under varied conditions.

To address these limitations, we propose \tool, an end-to-end automated and systematic framework for testing RAG-enabled LLMs. Our approach evaluates RAG systems across multiple qualitative dimensions, including factuality, relevance, completeness, and clarity. The proposed framework consists of four main steps. First, it automatically generates retrieval documents (we focus this paper on PDF files while remaining extensible to other formats). Second, it analyzes these documents to generate diverse test inputs along with their corresponding expected outputs. Third, it executes the generated queries on the RAG-LLM under test. Lastly, it employs an LLM-as-a-Judge mechanism to automatically assess whether the produced outputs match the expected ones.

We conduct a large-scale empirical evaluation involving eight state-of-the-art LLMs, including four closed-source and four open-weight models, combined with three different embedding techniques, resulting in 24 distinct configurations. Overall, we generated 3,000 test inputs and executed a total of 72,000 test cases. Our approach detected 21,633 failures, compared to 20,293 identified by a baseline variation or our tool, achieving a 6.6\% increase in failure detection. These results demonstrate that \tool is effective at uncovering subtle and diverse failure modes in RAG systems that are often overlooked by existing evaluation approaches.

\section{Background}

\subsection{Retrieval-Augmented Generation}

Retrieval-Augmented Generation (RAG) is an architectural paradigm that combines information retrieval with language generation to produce responses grounded in external knowledge~\cite{lewis2020rag}. Unlike standalone Large Language Models (LLMs), whose responses are primarily based on knowledge encoded during training, RAG-enabled LLMs retrieve information from an external corpus at inference time. This enables the incorporation of proprietary, domain-specific, or recently updated information without retraining the underlying model. By conditioning generation on retrieved evidence, RAG can improve the factual grounding, transparency, and traceability of generated responses~\cite{shuster2021retrieval,menick2022teaching}.

A typical RAG pipeline comprises several interconnected stages. First, the source documents are collected, preprocessed, and divided into smaller passages or \emph{chunks}. Each chunk is then transformed into a vector representation using an embedding model~\cite{karpukhin2020dense,reimers2019sentence}. These representations are stored in a vector database, which supports efficient similarity-based retrieval through implementations such as FAISS~\cite{johnson2019billion,douze2025faiss}.
At inference time, the user query is encoded using an embedding model compatible with the indexed documents. The resulting query vector is compared with the stored document vectors, and the top-$k$ most similar passages are selected as contextual evidence. These passages are incorporated into a prompt together with the original query, after which the generative model produces an answer conditioned on both sources of information. Depending on the specific architecture, the pipeline may also include query rewriting, metadata filtering, hybrid lexical--semantic retrieval, passage reranking, or context compression before generation.

The quality of the final response therefore depends on the correct operation of multiple components. The embedding model must preserve the semantic relationships between queries and document passages, the retrieval mechanism must identify the relevant evidence, and the prompt-construction strategy must present this evidence in a form that the generative model can effectively interpret. The model must then extract, combine, and express the retrieved information without introducing unsupported content.

Retrieval quality is particularly important because errors introduced during retrieval propagate to the generation stage~\cite{gao2023retrieval,izacard2023atlas}. When relevant passages are not retrieved, the model may lack the information required to answer the query. Conversely, the retrieval of irrelevant, incomplete, or contradictory passages may distract the model and lead to inaccurate or misleading responses. Nevertheless, successful retrieval does not guarantee a correct final answer: the generative model may ignore relevant evidence, misinterpret it, omit important details, or introduce information that is not supported by the retrieved context. Consequently, RAG systems must be considered end-to-end systems whose behavior emerges from the interaction among document processing, embeddings, retrieval, prompt construction, and language generation.

\subsection{Quality Assessment of RAG Systems}

The quality of a RAG system can be assessed at different levels. Component-level assessment evaluates individual stages of the pipeline, whereas end-to-end assessment examines the quality of the final response produced from a query and an external knowledge corpus. Both perspectives are necessary because the correctness of an individual component does not necessarily imply the correctness of the complete system.

The retrieval component is commonly assessed using information-retrieval metrics. Precision measures the proportion of retrieved passages that are relevant to the query, whereas recall measures the proportion of relevant passages that are successfully retrieved. Metrics such as Recall@$k$ evaluate whether relevant evidence appears among the top-$k$ retrieved results. Mean Reciprocal Rank measures how highly the first relevant result is ranked, while Normalized Discounted Cumulative Gain considers both the relevance and ordering of multiple retrieved passages. These metrics provide insight into the effectiveness of the embedding and ranking mechanisms, but they do not directly measure whether the generative model uses the retrieved evidence correctly.

End-to-end assessment focuses on the relationship among the query, the retrieved context, and the generated answer. Several complementary quality dimensions are commonly considered in RAG evaluation~\cite{gao2023survey,zhu2024rageval}. \emph{Answer relevance} captures whether the response directly addresses the user's query. \emph{Faithfulness}, also referred to as groundedness, assesses whether the claims contained in the response are supported by the retrieved evidence. \emph{Completeness} measures whether the response includes the essential information needed to answer the query, while \emph{clarity} concerns the coherence, precision, and readability of the generated text.

These dimensions capture different failure conditions. A response may be relevant but unfaithful when it directly answers the query using unsupported information. It may be faithful but incomplete when all included statements are supported by the context but important facts are omitted. Similarly, a system may retrieve the correct passage while producing an incorrect answer because the model fails to interpret or synthesize the available evidence. Assessing only final-answer accuracy may therefore obscure the origin and characteristics of failures within the RAG pipeline.

Grounding is especially important because retrieval augmentation reduces, but does not eliminate, hallucinations~\cite{shuster2021retrieval}. A generated response is grounded when its factual claims can be traced to the retrieved evidence. Evidence attribution mechanisms, such as requiring models to support answers with quotations or citations, can facilitate this assessment and improve the verifiability of generated information~\cite{menick2022verified}. However, a response may still contain subtle distortions, unsupported generalizations, or fabricated details even when it refers to the correct source material.

RAG evaluation is frequently conducted using question-answering datasets containing predefined queries and reference answers. Exact Match and token-level F1 are commonly used when answers are short and relatively constrained~\cite{lewis2020rag,izacard2023atlas}. However, these metrics are less suitable for open-ended natural-language responses because multiple semantically equivalent answers may differ substantially in wording. Consequently, semantic evaluation mechanisms are often required to distinguish acceptable linguistic variation from genuine behavioral failures.

\subsection{Testing and Test Oracles for RAG Systems}

From a software-testing perspective, a test case for a RAG system comprises at least a retrieval corpus, a test query, and an expected behavioral outcome. Depending on the testing objective, the expected outcome may specify a reference answer, a set of required facts, the passages that should be retrieved, or a rejection response when the corpus does not contain sufficient information.

Testing RAG systems is more complex than testing standalone LLMs because failures may originate from multiple interacting components. An incorrect response may result from inadequate document extraction, unsuitable chunk boundaries, inaccurate embeddings, retrieval errors, context truncation, ineffective prompt construction, or deficiencies in the generative model. Furthermore, the same query may produce different outcomes when the LLM, embedding model, retrieval configuration, or indexed corpus is changed. Effective testing must therefore consider both individual components and their end-to-end interaction.

An important distinction can be made between positive and negative test cases. Positive test cases ask questions whose answers are supported by the indexed documents and assess whether the system retrieves and uses the corresponding evidence. Negative test cases ask for information that is absent from the available corpus. In such cases, the desired behavior is generally to acknowledge that the available evidence is insufficient rather than generating a plausible but unsupported answer. Negative testing is therefore particularly useful for assessing hallucination resistance and the system's ability to respect the boundaries of its external knowledge.

Another relevant testing concept is coverage. In document-grounded systems, test coverage may refer to the extent to which different documents, sections, passages, topics, or linguistic structures contribute to the generated test cases. Generating many questions from a small or easily interpreted portion of a document may provide limited evidence about the behavior of the complete system. Coverage-oriented test generation instead seeks to exercise diverse parts of the retrieval corpus and a range of query characteristics, including questions that require direct extraction, synthesis of multiple facts, interpretation of complex passages, or rejection of unsupported requests.

Determining whether a RAG response is correct gives rise to a test oracle problem. Exact string comparison is often inadequate because natural-language answers may express the same information using different vocabulary, ordering, or levels of detail. Reference answers may also be incomplete, and a generated answer can contain additional correct information without necessarily constituting a failure.

A common strategy for addressing this problem is to employ an LLM as a semantic test oracle. In an LLM-as-a-judge configuration, a separate model compares the generated answer with a reference answer and, where available, the retrieved evidence. The judge can assess dimensions such as semantic equivalence, relevance, completeness, and faithfulness~\cite{zhu2024rageval}. This enables the automated evaluation of large numbers of open-ended responses without requiring exact lexical correspondence.

Nevertheless, LLM-based test oracles are themselves probabilistic and may produce incorrect or inconsistent verdicts. Their reliability depends on factors such as the selected judge model, the evaluation prompt, the information provided to the judge, and the complexity of the evaluated response. Consequently, the precision of an LLM-based oracle should be empirically assessed, particularly when it is used to classify system outputs as passing or failing.

Taken together, these characteristics show that testing a RAG system requires more than measuring retrieval accuracy or evaluating isolated model responses. It requires the systematic construction of retrieval corpora and queries, the exercise of both supported and unsupported information needs, the consideration of document and input coverage, and an oracle capable of evaluating semantically diverse natural-language outputs.

\section{\tool}

\subsection{Overview}

\begin{figure*}
	\centering
	\includegraphics[width=.9\textwidth]{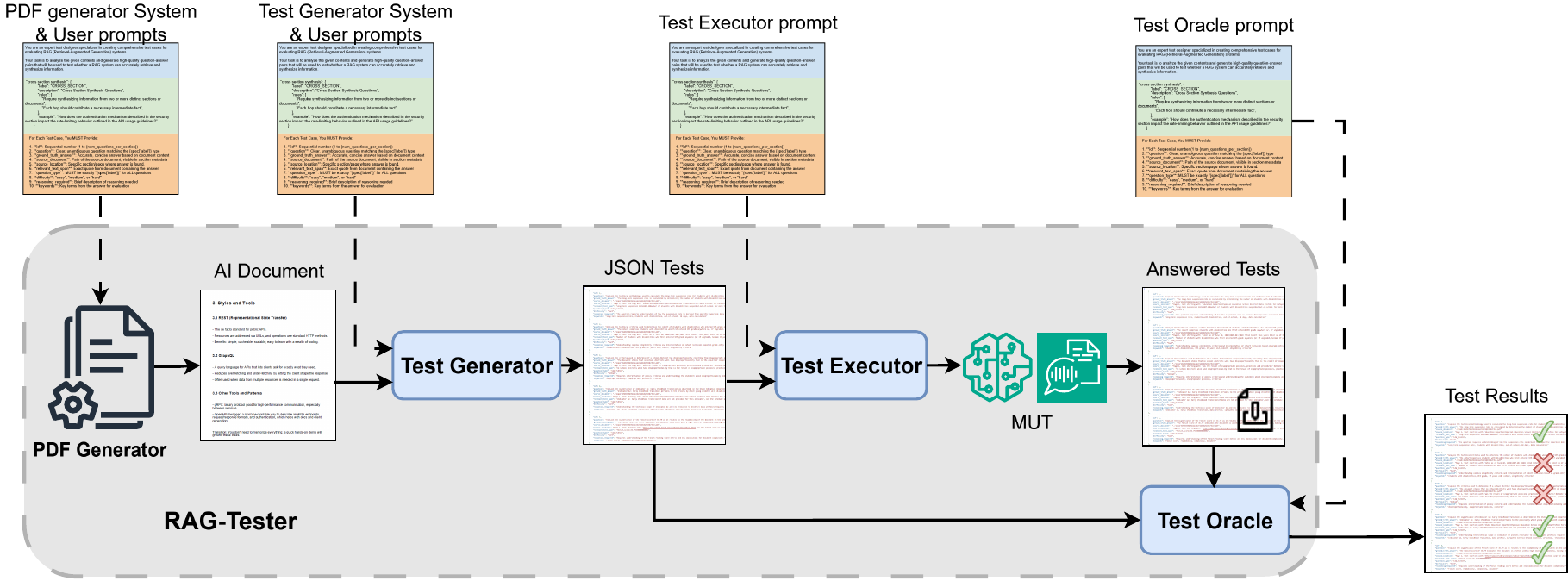}
	\caption{RAG-Tester execution flow architecture}
	\label{fig:main_architecture}
\end{figure*}

\tool consists of four main components that together form an end-to-end automated test execution framework for Retrieval-Augmented LLMs. Algorithm \ref{algo:RAGTester} and Figure \ref{fig:main_architecture} outline the main steps of \tool and illustrate how these four components interact. As input, \tool receives a text document and some instructions, producing a set of test results as output. The first task performed by \tool is the automated generation of test files (Line 2), which will later be retrieved by the LLM Under Test ($LLMUT$). To accomplish this, our tool uses an LLM to create files of varying lengths, complexities, and topics. These files are then merged (Line 3) with other documents in $D_{man}$, which test engineers may optionally include manually. This enables, among other benefits, the evaluation of RAG systems in specific application contexts. In the second phase, \tool processes the documents in $D$ and generates pairs of test inputs ($TI$) and expected test outputs ($TO_{exp}$) (Line 4). The number of generated test cases for each document in $D$ is $N_{test}$. These test inputs are then executed in batch (Line 5), and the resulting outputs ($TO$) are evaluated by comparing them with the expected outputs using our test oracle (Line 6), which provides the test results.

	\setlength{\algomargin}{4mm}
    
	\begin{algorithm}[h!]
    
	\DontPrintSemicolon
	 \SetKwBlock{DoParallel}{do in parallel}{end}
	\linespread{0.8}\selectfont
	\footnotesize
	\SetAlCapHSkip{0em}
	\SetKwProg{Fn}{function}{}{}
	\SetKwInOut{Input}{input}
	\SetKwInOut{Output}{output}

\Input{$N_f$: Number of files to be retrieved\\ 
$D_{man}$ =\{$d_1,d_2, ..., d_m$\}: Manually included files to be retrieved \\
$N_{test}$: Number of test cases to generate for each retrieval document\\

}

\Output{$TR$: Final test results}
\BlankLine
\Fn{\textsc{\textbf{RAGTester}}}{
         \textit{$D$} $\leftarrow$ \textsc{GenerateFiles($N_f$)}\;
        \textit{$D$} $\leftarrow$ $D\bigcup$ $D_{man}$\;
        $TI,TO_{exp}$ $\leftarrow$ \textsc{GenerateTestInputs($D, N_{test}$)}\;
        $TO$ $\leftarrow$ \textsc{ExecuteTestInputs($LLMUT, TI$)}\;
        $TR$ $\leftarrow$ \textsc{EvaluateTests($TO,TO_{exp}$)}
        
\Return{$TR$ } 
}

	\caption{{\tool's overall overview}}
    \label{algo:RAGTester}
\end{algorithm}

\subsection{Test File Generator}



Our initial prototype focuses on generating PDF files to be retrieved by the $LLMUT$, although \tool can be easily extended to support other file types. \tool leverages an LLM, specifically GPT-5 Nano, to automatically produce PDF files of varying sizes, levels of complexity, and topics. Designing effective system and user prompts is paramount for guiding this generation process. Appendices \ref{appendix:pdf_system} and \ref{appendix:pdf_user} show the employed prompts for this LLM.

On the one hand, we specify the technical aspects of the document in the \textit{System Prompt}, these being the desired length, level of detail (e.g., comprehensive and detailed, concise summary, highly technical...), and the tone or writing style (e.g., formal or casual). We further define elements related to document structure, content requirements, and formatting guidelines. These configurations prepare the LLM for the subsequent user prompt. On the other hand, in the \textit{User Prompt}, the topic and name of the document is provided optionally, although the LLM can autonomously define both. The user can also provide information on whether the document should contain any specific section.

The LLM responsible for generating the retrievable file produces its output in Markdown format. The file generator then automatically converts this Markdown content into a PDF file.





\subsection{Test Input Generator}

After generating the files to be retrieved by the $LLMUT$, \tool focuses on the generation of the test inputs ($TI$) and expected test outputs ($TO_{exp}$). For each of the files in $D$, the test input generator generates a set of $N_{test}$ test cases. 

It is worth mentioning that each tested source document is divided into $N_{test}$ sections (depending on the amount of specified queries), each of these being delimited by either a paragraph end or a maximum amount of tokens. After the splitting is done, each chunk is used for one test query only, ensuring that no text region is used multiple times.

To generate these queries, a system and user prompt strategy is used, similar to test file generation. The system prompt (Appendix \ref{appendix:testgen_system}) is used to determine the technical instructions, such as question type specifications—these being the label to include in each test, short description, and a set of rules to follow when generating—required output format (i.e., a list of the information the LLM needs to provide) and quality standards, so the queries are clear, test different aspects of the document, and cannot be answered with only \textit{yes} or \textit{no}. The user prompt (Appendix~\ref{appendix:testgen_user}) specifies the document chunks to be considered and the number of queries to generate. When the selected question type is Natural Language Processing, the chunks may be ordered according to their Flesch Reading Ease score. Specifically, the generator first measures the complexity of each document section using the Flesch Reading Ease score~\cite{flesch1948new}, a well-established readability metric ranging from 0 to 100, where lower values indicate greater textual complexity. The sections are then sorted in ascending order by their scores, and the test input generator selects the $N_{complex}$ most complex paragraphs, where $N_{complex}<N_{test}$. GPT-4.1 Nano is subsequently used to generate a corresponding question for each selected paragraph. We prioritize complex passages under the assumption that they are more difficult for the $LLMUT$ to interpret, thereby increasing the likelihood that \tool exposes failures.


Secondly, \tool's test input generation strategy also produces a set of negative rejection-type queries. These questions are intentionally crafted around information that is not present in the source documents. Their purpose is to verify whether the $LLMUT$ appropriately rejects unsupported claims instead of hallucinating answers. To generate these queries, \tool uses two complementary methods. 
The first method aims at not including sections that are available in the document, asking the LLM to generate questions randomly. The second method chooses a different document from
$D$, and repeating the same chunking process mentioned previously, ensuring that the generated question will have no relationship with the document originally chosen for retrieval. These two approaches allow \tool to assess the model's robustness in handling absent or incomplete evidence.

Lastly, our strategy incorporates a coverage-based mechanism aimed at maximizing the portions of the document that remain non-generated by the previous two steps. While complexity-based and rejection-type queries target specific types of weaknesses, they may not uniformly cover all sections of the input documents. To address this, the test input generator analyzes which paragraphs or content areas have not yet contributed to any test case and selects additional segments to ensure broader coverage. By prompting an LLM to generate questions from these underrepresented parts of the document, \tool increases the diversity of test inputs and enhances the likelihood of exposing failures that might arise from overlooked or less prominent content.


\subsection{Test Executor}


Once the test case files are generated for all selected documents, these test queries can be answered using different Model Under Test combinations, each of these being formed by an LLM and an embedding model. Some of the essential components for the test executor include: the RAG chain, vector stores, and the prompt template.

Regarding the embeddings, Chroma vector stores are used. This way, each document can have its dedicated embedding store, and since the documents are being used repetitively, once a vector store is created it eliminates the need to create a new one, making it more efficient both computationally and time wise. The embeddings, RAG chain and prompt template are created using LangChain~\cite{langchain2023}. The RAG chain is made combining the LLM and the prompt template, the first one being either OpenAI or Ollama sourced, and the prompt template specifies the the role of the model and what instructions to follow, for it is told that if it does not have enough information to answer the test appropriately, it should output \textit{"Insufficient information to answer this question"} or \textit{"I don't know"}. Additionally, it also includes the question and the document sections, as well as how to format the generated output.

The flow of the executions starts by matching the test files to their original documents, so in each iteration the information is only taken from the source text file. After electing the document, all test queries are executed by all $LLMUT$, outputting the results in a JSON format and parsing them to a JSON file. Each $LLMUT$ and $D$ create one answer file.


\subsection{Test Oracle}

The last phase of our approach is the evaluation. To this end, we employ an LLM-as-a-judge strategy to solve the well-known test oracle problem in the context of testing LLMs. While simple, this strategy is also employed by other LLM testing techniques(e.g.,~\cite{zhu2024rageval,romero2026meta}). The evaluator obtains a set of Test Outputs ($TO$) provided by the $LLMUT$ and a set of Expected Test Outputs ($TO_{exp}$) generated by the test input generator. The oracle inspects one by one  whether both outputs are semantically equivalent. The specific prompt used for this task is shown in Appendix \ref{appendix:oracle_prompt}. 
Based on preliminary analyses, we selected GPT-4.1 Nano as the LLM-as-a-judge of \tool, although our approach can easily integrate any other LLM. It is important to note that it is convenient that this LLM is different from the one being under test. 
Given both $TO$ and $TO_{exp}$, the LLM-as-a-judge assesses the following four criteria:
\begin{itemize}
  \item \textbf{Faithfulness}: It measures the degree to which the model’s answer is factually grounded in, and semantically consistent with, the expected answer and the retrieved evidence. A response is considered faithful if it does not introduce unsupported claims, contradictions, or hallucinated information beyond what is justified by the source document. This criterion is particularly critical in RAG systems, where generation must remain strictly aligned with retrieved context. Any deviation from the factual content of the ground truth, including subtle distortions or fabricated details, negatively impacts the faithfulness score. In our framework, faithfulness acts as a primary correctness condition: if it is evaluated as false, the overall verdict is automatically Failed.
  \item \textbf{Relevance}: It evaluates whether the generated response directly addresses the query posed in the test input. A response may be factually correct yet partially irrelevant if it includes tangential information, digressions, or extraneous explanations that do not contribute to answering the specific question. The relevance criterion therefore focuses on alignment between the query intent and the produced answer. High relevance indicates that the model remains focused on the task and avoids unnecessary content generation.
  \item \textbf{Completeness}: It assesses the extent to which the answer covers all essential elements present in the expected output. In many RAG scenarios, correct responses require the inclusion of multiple facts, conditions, or relationships extracted from the source document. An answer that omits critical components is considered partially complete, even if the included information is correct. Importantly, if the Model Under Test provides additional correct and supported details beyond those in the reference answer, this is evaluated positively, as long as faithfulness is preserved. Completeness therefore captures the coverage of required information rather than strict equivalence in wording.
  \item \textbf{Clarity}: It examines the linguistic and structural quality of the response. This includes coherence, logical organization, grammatical correctness, and precision of language. Even when a response is factually correct, poor organization or ambiguous phrasing may reduce its practical utility. Clarity ensures that answers are not only correct but also understandable and well-structured. This criterion is especially relevant in real-world RAG applications, where outputs are consumed directly by end users.
  \item \textbf{Keywords match}: It provides an additional lexical-level verification mechanism. During test generation, each expected answer is associated with a set of essential keywords representing critical entities, concepts, or terms that should appear in a correct response. The oracle checks whether these keywords are present in the model’s answer. A rating of true indicates that all required keywords appear, partially indicates that only a subset is present, and false indicates that none appear. While keyword matching alone is insufficient to determine semantic correctness, it serves as a complementary signal that reinforces completeness and faithfulness assessments.
\end{itemize}


Each criterion is assigned one of three possible ratings: \textbf{True}, \textbf{Partially} or \textbf{False}, which for scoring purposes are mapped to numerical values \textbf{1}, \textbf{0.5} and \textbf{0}, respectively. The overall verdict is calculated by taking the arithmetic mean of the five criterion scores. If the final average is below 0.5, the answer receives a \textbf{Fail} verdict. If the average is 0.5 or higher, the answer receives a \textbf{Pass} verdict. It is also important to note that if the criteria Faithfulness is False, we consider the test as failed.


\section{Tool Support}

We provide \tool as an Open-Source end-to-end RAG-LLM test generation tool that enables testing Retrieval Augmented Language Models via a user-friendly web interface or a code-based environment. Both interfaces provide parity in core functionalities: PDF generation, test generation, test execution and test evaluation.

While the code-based version allows for direct execution and debugging, the web interface offers a more appealing test result visualization, making it easier to identify strengths and weaknesses. Furthermore, \tool is engineered with a modular architecture, so if developers wish to implement new LLM providers, document loaders, etc. the process remains simple and extensible. A demo for the tool using the web interface can be found online \cite{ragtester_demo}.

\section{Experimental Design}

\subsection{Research Questions}

Our evaluation aimed at answering the following four Research Questions (RQs):

\begin{itemize}
    \item \textit{\textbf{RQ1 -- Overall performance:} How does \tool perform at detecting failures on RAG-enabled LLMs?} With this RQ we wanted to assess the overall performance provided by \tool when testing different LLM and embedding combinations.

    \item \textit{\textbf{RQ2 -- Comparison with Baseline:} How does \tool compare with a baseline test input generator?} In this RQ we aimed at comparing \tool with a simplified version of the test input generator we developed.

    \item \textit{\textbf{RQ3 -- PDF Generation Strategies:} Which is the difference between generating the context PDF files automatically or employing existing PDF files?} In the third RQ we aimed at answering whether there are differences in generating the context files in an automated manner. 

    \item \textit{\textbf{RQ4 -- Oracle Precision:} Which is the precision of the test oracles of \tool?} Since \tool's oracle employs an LLM-as-a-judge, it might have certain proneness to hallucination. This RQ aims at checking its precision to investigate whether and to which extent the proposed test oracle triggers false positives. 
\end{itemize}

\subsection{Evaluation Metrics}

To evaluate the performance of the Test Oracle, a sample of 382 failed tests was drawn from a total of 41,926 detected failures (combining all baseline and \tool tests, on both OpenAI and Ollama) using Cochran's Sample Size Formula. The goal was to inspect this sample and classify each test into one of two categories:

\begin{itemize}
    \item \textbf{True Positives (TP)}: A failed test correctly identified as FAIL
    \item \textbf{False Positives (FP)}: A passed test incorrectly identified as FAIL
\end{itemize}

Though not used, the remaining two categories were also defined:

\begin{itemize}
    \item \textbf{False Negatives (FN)}: A failed test incorrectly identified as PASS
    \item \textbf{True Negatives (TN)}: A passed test correctly identified as PASS
\end{itemize}

From these classifications, two key performance metrics were then calculated: Oracle Precision, which measures the proportion of tests flagged as failures that are genuinely failing, and the False Discovery Rate (FDR), which captures the proportion that are incorrectly flagged. Together, these metrics provide a clear picture of how reliably the Test Oracle distinguishes true failures from false alarms.

Since manually reviewing all 382 tests would be extremely time-consuming, a majority voting strategy was used instead. Four LLMs larger than the Test Oracle (GPT-4o, GPT-5, Llama 3.1 70B and Qwen 2.5 72B), each independently evaluated every test, and the most common verdict across the four models was selected as the final classification.

\subsection{Baseline Technique}

For the baseline, we removed some of the key features of the original test input generation algorithm.
Specifically, the baseline gets the context document (i.e., a PDF) and generates $N_{test}$ number of test inputs by not following any strategy (e.g., without considering document coverage nor paragraph complexity). 

\subsection{Test generation Setup}

In our evaluation, we automatically generated 30 PDF files and we manually selected 30 manually selected PDF files from Kaggle \cite{manisha717_pdf_dataset}, i.e., we employed a total of 60 PDF files as context. For each of the documents, 50 different test inputs were generated by each of the techniques, i.e., \tool and the baseline. That is, \tool and its baseline version generated 3,000 test cases each. 

The rationale for selecting the 30 PDFs from Kaggle were the following: Each document had to be at least 5 pages long, and the selection as a whole needed to cover a diverse range of formats, including documents with tables and images as well as simpler text-only files. Documents containing headers, footers and graphical elements were prioritized, in addition to those with varied text styles and font sizes. This diversity was intentional, as testing across different document structures gives a more thorough picture of how well the models handle real-world PDFs.

\subsection{Selected Large Language Models and Embeddings}

We selected both, closed-sourced and open-weight LLMs for being used as testing subjects. Specifically, we employed four models of OpenAI (i.e., GPT-3.5 Turbo, GPT-4.1 Mini, GPT-4o Mini and GPT-5 Nano) and four open-weight models (i.e., Gemma 3:8b, Qwen 3:8b, DeepSeek R1:8b and Llama 3.2:3b). The criteria for the selection of the closed-source models were (1) they are widely used and form state-of-the-art LLMs and (2) their token costs is not excessive. On the other hand, for the open-weight models, we employed widely used LLMs, but that were executable in our GPU clusters on top of Ollama. Higher parameter LLMs were not executable due to their high computational cost.

In addition, different text embeddings were selected. For OpenAI's models, OpenAI provided the option for using these three different embeddings: text-embedding-3-small, text-embedding-3-large, text-embedding-ada-002. Meanwhile, for the open-weight LLMs, we used three other compatible embeddings: bge-large:335m-en-v1.5-fp16, nomic-embed-text:v1.5 and intfloat-e5-base-v2:q8\_0. Note that those embeddings from OpenAI's were not compatible with Ollama.

\section{Analysis of the Results and Discussion}

\subsection{RQ1 -- Overall Effectiveness}

\tool proved highly effective at detecting failures in RAG-enabled LLMs across all 24 tested combinations (4 LLMs x 3 embedding models in 2 LLM providers).
Table \ref{tab:openai_mut_failures} presents the failure counts for the four OpenAI models (GPT-3.5-Turbo, GPT-4.1-Mini, GPT-4o-Mini, and GPT-5 Nano) paired with the three OpenAI embedding models (text-embedding-3-small, text-embedding-3-large, and text-embedding-ada-002). Table \ref{tab:ollama_mut_failures} shows the equivalent results for the four open-weight models (Gemma-3:8b, Qwen-3:8b, DeepSeek-R1:8b, and Llama-3.2:3b) paired with the three compatible embeddings (bge-large:335m-en-v1.5-fp16, nomic-embed-text:v1.5, and intfloat-e5-base-v2:q8\_0).

Across \tool's 72,000 total test executions (3,000 queries per technique × 24 combinations, executed on the 60 context PDFs),
\tool consistently surfaced failures stemming from retrieval inaccuracies, incomplete context integration, hallucinated content on negative queries, and poor handling of complex passages. Specifically, \tool triggered a total of 8,880 failures in OpenAI's LLMs, and 12,753 failures in open-weight models. Failure rates varied by model family and embedding quality, with lower-parameter open-weight models and smaller embeddings exhibiting noticeably higher failure counts. These results confirm that \tool successfully exposes subtle failure modes, validating its utility for systematic RAG-LLM evaluation.

These results demonstrate that \tool is highly effective at surfacing subtle, real-world RAG weaknesses that traditional benchmarks (e.g., exact-match on Natural Questions) miss entirely. Even the strongest configuration (GPT-4.1 Mini + text-embedding-3-large) still failed 16,93\% of the time, showing that ``state-of-the-art'' does not necessarily mean ``reliable RAG''. The wide performance gap between closed-source and open-weight models (OpenAI group averaged 24.67\% failures vs. 35.43\% for Ollama) also highlights the current maturity difference between proprietary and open ecosystems. Practically, this means developers should never deploy a RAG system without running a systematic test suite like \tool, especially in high-stakes domains (law, medicine, finance) where hallucination or incomplete answers can have serious consequences. 

Some examples of failure types we discovered were related to (1) hallucinations, (2) retrieval inaccuracies, and (3) comprehension deficits.

These failures were most prominent across the previously mentioned test query types:

\begin{itemize}
    \item \textbf{Negative rejection}: Instances where the model hallucinated a plausible-sounding but unsupported answer as seen in Table \ref{tab:ragtester_negative_rejection}, failing to provide a clear rejection signal when the necessary knowledge is absent from all retrieved text sections.

    \item \textbf{Document maximizing}: "Generic" queries that require synthesized data from across the entire text, which often resulted in incorrectly retrieved context when the system failed to aggregate multiple relevant paragraphs, as illustrated in Table \ref{tab:ragtester_document_maximizing}.

    \item \textbf{Natural Language Processing} (Low Flesch Score): Tests using highly complex, dense text sections where the model frequently exhibited failed generation due to an inability to parse nuanced technical structures, as reported in Table \ref{tab:ragtester_nlp}.
\end{itemize}

\begin{table*}[t]
\centering
\small
\setlength{\tabcolsep}{6pt}
\renewcommand{\arraystretch}{1.2}
\caption{RAG-Tester negative rejection example}

\resizebox{\textwidth}{!}{
    \begin{tabular}{cllll}
\multicolumn{5}{c}{{\color[HTML]{000000} \textbf{Negative Rejection (NEG\_REJ)}}}                                                                                 \\ \hline
\multicolumn{1}{c|}{\cellcolor[HTML]{FFFFFF}\textbf{Test Query}}                                                                                                 & \multicolumn{1}{c|}{\cellcolor[HTML]{FFFFFF}\textbf{Answer Generated by}}                                                                                                                                                                                                        & \multicolumn{1}{l|}{\cellcolor[HTML]{FFFFFF}\textbf{Ground Truth Answer}}                                                                 & \multicolumn{1}{l|}{\textbf{\begin{tabular}[c]{@{}l@{}}Retrieved text by\\ Embedding UT\end{tabular}}}                                                                                                              & \textbf{\begin{tabular}[c]{@{}l@{}}Retrieved text used\\ to generate Ground\\ Truth Answer\end{tabular}} \\ \hline
\multicolumn{1}{l|}{\cellcolor[HTML]{FFFFFF}\begin{tabular}[c]{@{}l@{}}What are the three\\ primary functions of\\ the human respiratory\\ system?\end{tabular}} & \multicolumn{1}{l|}{\cellcolor[HTML]{FFFFFF}\begin{tabular}[c]{@{}l@{}}The three primary\\ functions of the\\ human respiratory\\ system are to deliver\\ oxygen to the body,\\ remove carbon dioxide\\ from the body, and\\ help regulate the\\ body’s pH levels.\end{tabular}} & \multicolumn{1}{l|}{\cellcolor[HTML]{FFFFFF}\begin{tabular}[c]{@{}l@{}}Insufficient\\ information to answer\\ this question\end{tabular}} & \multicolumn{1}{l|}{\begin{tabular}[c]{@{}l@{}}The afternoon session\\ broadened the focus\\ to the question of\\ how to integrate\\ restoration into\\ managing watersheds\\ for flood protection...\end{tabular}} & unknown                                                                                                 
\end{tabular}

}
\label{tab:ragtester_negative_rejection}
\end{table*}

\begin{table*}[t]
\centering
\small
\setlength{\tabcolsep}{6pt}
\renewcommand{\arraystretch}{1.2}
\caption{RAG-Tester document maximizing example}

\resizebox{\textwidth}{!}{
    \begin{tabular}{cllll}
\multicolumn{5}{c}{{\color[HTML]{000000} \textbf{Document Maximizing (DOC\_MAX)}}}\\ \hline
\multicolumn{1}{c|}{\cellcolor[HTML]{FFFFFF}\textbf{Test Query}}                                                                                                         & \multicolumn{1}{c|}{\cellcolor[HTML]{FFFFFF}\textbf{Answer Generated by}}                                                                                                                                                                                                                                     & \multicolumn{1}{l|}{\cellcolor[HTML]{FFFFFF}\textbf{Ground Truth Answer}}                                                                                                                                                                   & \multicolumn{1}{l|}{\textbf{\begin{tabular}[c]{@{}l@{}}Retrieved text by\\ Embedding UT\end{tabular}}}                                                                                                                                                                                      & \textbf{\begin{tabular}[c]{@{}l@{}}Retrieved text used\\ to generate Ground\\ Truth Answer\end{tabular}}                                                                                                                                                                         \\ \hline
\multicolumn{1}{l|}{\cellcolor[HTML]{FFFFFF}\begin{tabular}[c]{@{}l@{}}What are the common\\ metrics used for\\ evaluating continuous\\ prediction models?\end{tabular}} & \multicolumn{1}{l|}{\cellcolor[HTML]{FFFFFF}\begin{tabular}[c]{@{}l@{}}Common metrics used\\ for evaluating\\ continuous prediction\\ models include Mean\\ Squared Error (MSE),\\ Pseudo-Earth Mover\\ Divergence (pEMDiv),\\ Brier score, and\\ Continuous risk\\ probability score\\ (CRPS).\end{tabular}} & \multicolumn{1}{l|}{\cellcolor[HTML]{FFFFFF}\begin{tabular}[c]{@{}l@{}}Mean Squared Error\\ (MSE), Pseudo-Earth\\ Mover Divergence\\ (pEMDiv), and\\ Targeted Absolute\\ Difference with\\ Direction\\ Augmentation\\ (TADDA)\end{tabular}} & \multicolumn{1}{l|}{\begin{tabular}[c]{@{}l@{}}Metrics for continuous\\ prediction models\\ include: Mean\\ Squared Error (MSE):\\ Common and simple\\ measure of how close\\ the predicted value is\\ to the observed value.\\ Pseudo-Earth Mover\\ Divergence\\ (pEMDiv)...\end{tabular}} & \begin{tabular}[c]{@{}l@{}}Metrics for continuous\\ prediction models\\ include: ˘2022 Mean\\ Squared Error (MSE):\\ ... ˘2022 Pseudo-Earth\\ Mover Divergence\\ (pEMDiv): ... ˘2022\\ Targeted Absolute\\ Difference with\\ Direction\\ Augmentation\\ (TADDA):...\end{tabular}
\end{tabular}
}
\label{tab:ragtester_document_maximizing}
\end{table*}

\begin{table*}[t]
\centering
\small
\setlength{\tabcolsep}{6pt}
\renewcommand{\arraystretch}{1.2}
\caption{RAG-Tester natural language processing example}

\resizebox{\textwidth}{!}{
    \begin{tabular}{cllll}
\multicolumn{5}{c}{{\color[HTML]{000000} \textbf{Natural Language Processing (LOW\_FLESCH)}}}\\ \hline
\multicolumn{1}{c|}{\cellcolor[HTML]{FFFFFF}\textbf{Test Query}}                                                                                                                                              & \multicolumn{1}{c|}{\cellcolor[HTML]{FFFFFF}\textbf{Answer Generated by}}                                                                                                                                                                                                               & \multicolumn{1}{l|}{\cellcolor[HTML]{FFFFFF}\textbf{Ground Truth Answer}}                                                                                                                                                                               & \multicolumn{1}{l|}{\textbf{\begin{tabular}[c]{@{}l@{}}Retrieved text by\\ Embedding UT\end{tabular}}}                                                                                                                                                                                                                                   & \textbf{\begin{tabular}[c]{@{}l@{}}Retrieved text used\\ to generate Ground\\ Truth Answer\end{tabular}}                                                                                                                                                                                    \\ \hline
\multicolumn{1}{l|}{\cellcolor[HTML]{FFFFFF}\begin{tabular}[c]{@{}l@{}}What are the key\\ financial figures\\ associated with New\\ Mexico’s conservation\\ efforts and programs\\ as reported?\end{tabular}} & \multicolumn{1}{l|}{\cellcolor[HTML]{FFFFFF}\begin{tabular}[c]{@{}l@{}}The key financial\\ figures associated\\ with New Mexico’s\\ conservation efforts\\ are \$5,098,856 and\\ \$6,632,204 for\\ livestock water\\ development and\\ alleviation of brush,\\ respectively\end{tabular}} & \multicolumn{1}{l|}{\cellcolor[HTML]{FFFFFF}\begin{tabular}[c]{@{}l@{}}The 2002 Farm Bill\\ invested over \$6.2\\ million in projects,\\ and NRCS received\\ nearly \$31 million for\\ the EQIP program,\\ benefiting thousands\\ of acres.\end{tabular}} & \multicolumn{1}{l|}{\begin{tabular}[c]{@{}l@{}}Seventeen percent of\\ the Farm Bill funds in\\ New Mexico have\\ gone to livestock\\ water development,\\ amounting to an\\ investment of\\ \$5,098,856. With\\ these funds, ranchers\\ have installed 450\\ wells, 1,179 miles of\\ pipeline, and 1,314\\ watering tanks.\end{tabular}} & \begin{tabular}[c]{@{}l@{}}May 13, 2003 marked\\ the first anniversary of\\ the 2002 Farm Bill.\\ The 2002 Farm Bill\\ has already invested\\ \$6,267,457 in 151\\ projects in New\\ Mexico... Since that\\ time, NRCS has\\ received \$30,883,345\\ for its EQIP\\ program 2026\end{tabular}
\end{tabular}
}
\label{tab:ragtester_nlp}
\end{table*}

\begin{table*}[t]
\centering
\small
\setlength{\tabcolsep}{6pt}
\renewcommand{\arraystretch}{1.2}
\caption{Baseline retrieval inaccuracy example}

\resizebox{\textwidth}{!}{
    \begin{tabular}{cllll}
\multicolumn{5}{c}{{\color[HTML]{000000} \textbf{Retrieval Inaccuracy}}}\\
\hline
\multicolumn{1}{c|}{\cellcolor[HTML]{FFFFFF}\textbf{Test Query}}                                                                                                                            & \multicolumn{1}{c|}{\cellcolor[HTML]{FFFFFF}\textbf{Answer Generated by}}                                                                                                                   & \multicolumn{1}{l|}{\cellcolor[HTML]{FFFFFF}\textbf{Ground Truth Answer}}                                                                                                                                                         & \multicolumn{1}{l|}{\textbf{\begin{tabular}[c]{@{}l@{}}Retrieved text by\\ Embedding UT\end{tabular}}}                                                                                                                                                                                                      & \textbf{\begin{tabular}[c]{@{}l@{}}Retrieved text used\\ to generate Ground\\ Truth Answer\end{tabular}}                                        \\ \hline
\multicolumn{1}{l|}{\cellcolor[HTML]{FFFFFF}\begin{tabular}[c]{@{}l@{}}What is a key\\ recommendation for\\ making conflict\\ models more useful\\ for anticipatory\\ action?\end{tabular}} & \multicolumn{1}{l|}{\cellcolor[HTML]{FFFFFF}\begin{tabular}[c]{@{}l@{}}Utilize flexible models\\ that do not\\ pre-suppose a\\ theoretical framework\\ of conflict causality.\end{tabular}} & \multicolumn{1}{l|}{\cellcolor[HTML]{FFFFFF}\begin{tabular}[c]{@{}l@{}}Focus on predicting\\ conflict shifts\\ (escalation or onset)\\ rather than exact\\ conflict events, using\\ flexible and ensemble\\ models.\end{tabular}} & \multicolumn{1}{l|}{\begin{tabular}[c]{@{}l@{}}To make use of\\ conflict prediction for\\ anticipatory action in\\ the humanitarian\\ sector, we recommend\\ that future work:\\ ˘2022 Utilize flexible\\ models that do not\\ pre-suppose a\\ theoretical framework\\ of conflict causality.\end{tabular}} & \begin{tabular}[c]{@{}l@{}}Models should focus\\ on predicting shifts in\\ conflicts, such as\\ increases in intensity\\ or onset.\end{tabular}
\end{tabular}
}
\label{tab:baseline_retrieval_flaw}
\end{table*}

\begin{table*}[t]
\centering
\small
\setlength{\tabcolsep}{6pt}
\renewcommand{\arraystretch}{1.2}
\caption{Baseline generation flaw example}

\resizebox{\textwidth}{!}{
    \begin{tabular}{cllll}
\multicolumn{5}{c}{{\color[HTML]{000000} \textbf{Generation Flaw}}}\\
\hline
\multicolumn{1}{c|}{\cellcolor[HTML]{FFFFFF}\textbf{Test Query}}                                                                                                                    & \multicolumn{1}{c|}{\cellcolor[HTML]{FFFFFF}\textbf{Answer Generated by}}                                                                                                                                            & \multicolumn{1}{l|}{\cellcolor[HTML]{FFFFFF}\textbf{Ground Truth Answer}}                                                                                                                                                                                                  & \multicolumn{1}{l|}{\textbf{\begin{tabular}[c]{@{}l@{}}Retrieved text by\\ Embedding UT\end{tabular}}}                                                                                                                                   & \textbf{\begin{tabular}[c]{@{}l@{}}Retrieved text used\\ to generate Ground\\ Truth Answer\end{tabular}}                                                                                                             \\ \hline
\multicolumn{1}{l|}{\cellcolor[HTML]{FFFFFF}\begin{tabular}[c]{@{}l@{}}What is meant by\\ ’hitting the sector’s\\ water networks’ as an\\ example of cyber\\ threats?\end{tabular}} & \multicolumn{1}{l|}{\cellcolor[HTML]{FFFFFF}\begin{tabular}[c]{@{}l@{}}A malware operation\\ targeting the water\\ networks of a specific\\ sector to compromise\\ their security and\\ functionality.\end{tabular}} & \multicolumn{1}{l|}{\cellcolor[HTML]{FFFFFF}\begin{tabular}[c]{@{}l@{}}It refers to\\ cyber-attacks\\ targeting water\\ infrastructure, such as\\ the incident reported\\ for Israel’s Mekorot\\ water provider in\\ 2021, risking\\ operational disruption.\end{tabular}} & \multicolumn{1}{l|}{\begin{tabular}[c]{@{}l@{}}srael’s national water\\ provider Mekorot\\ reported in 2021 that\\ its water networks are\\ subject to ’several\\ hundreds of thousands\\ of hacking attempts’\\ per year.\end{tabular}} & \begin{tabular}[c]{@{}l@{}}Israel’s national water\\ provider Mekorot\\ reported in 2021 that\\ its water networks are\\ subject to ’several\\ hundreds of thousands\\ of hacking attempts’\\ per year.\end{tabular}
\end{tabular}
}
\label{tab:baseline_generation_flaw}
\end{table*}

Based on these results, the first RQ can be answered as follows: 

\begin{center}
\fbox{
\begin{minipage}{0.94\linewidth}
\textbf{Answer to RQ1.}
\tool effectively detected failures across all 24 combinations of LLMs and embedding models, identifying 21,633 failures in 72,000 test executions. The generated tests exposed heterogeneous failure modes, including retrieval inaccuracies, unsupported hallucinations, incomplete context integration, and difficulties interpreting complex passages. These results demonstrate the ability of \tool to systematically reveal weaknesses throughout the RAG pipeline.
\end{minipage}
}
\end{center}

\subsection{RQ2 -- Comparison with Baseline}

For the 24 different models and embedding combinations, \tool outperformed the baseline in 20 of the cases. Overall, \tool detected 6.22\% more failures than the baseline (21,633 vs. 20,293). For closed loop models (Table \ref{tab:openai_mut_failures}), \tool found 8,880 failures vs. the baseline’s 7,132 (+19.68\%). Conversely, in open-weight models, the baseline detected slightly more failures (13,161 vs. 12,753), but this was caused by the Llama 3.2 model, which was found especially weak. In this specific model, the baseline outperformed \tool, whereas in the rest  of the models(with the exception of Gemma 3 and the intfloat-e5-base-v2 embedding), \tool performed better.

These results demonstrate that the advanced test-input generation strategy in \tool (Flesch-based complexity sorting, full-document coverage guarantees, and targeted negative-rejection synthesis) is particularly powerful on stronger, production-grade models. On closed-source LLMs the structured approach uncovers substantially more real failure modes that a naive random baseline misses. The three cases where the baseline performed better (Llama-3.2 across all embeddings) are explained by the model’s extreme weakness: its overall failure rate exceeded 50\%, so even random queries frequently triggered detectable errors. The single Gemma-3 exception with intfloat-e5-base-v2 is minor and does not change the overall trend. From a practical standpoint, the 19.68\% improvement on closed-source models is highly significant, as organizations using GPT-class models would severely underestimate risk with a simple baseline. These findings validate the need for researching advanced test generation strategies for RAG-enabled LLMs, for which we strongly recommend using the full \tool generator rather than any simpler baseline when reliability is paramount.

In comparison to \tool, the discovered failure types on the baseline were related to (1) retrieval inaccuracies and (2) incorrectly generated answer. No different test query types were used in the baseline, so no hallucination failures were detected. The detected failures can be described as such:

\begin{itemize}
    \item \textbf{Retrieval inaccuracies}: When the retriever pulls text that is irrelevant, partially relevant but incomplete, or entirely noise (unrelated to query), as seen in Table \ref{tab:baseline_retrieval_flaw}.
    
    \item \textbf{Incorrectly generated answer}: The model fails to extract, reason through, shown in Table \ref{tab:baseline_generation_flaw}, or correctly format the answer even if the correct source material is available in the prompt.
\end{itemize}

In summary, RQ2 can be answered as follows:

\begin{center}
\fbox{
\begin{minipage}{0.94\linewidth}
\textbf{Answer to RQ2.}
\tool outperformed the baseline in 20 of the 24 evaluated LLM--embedding configurations and detected 21,633 failures, compared with 20,293 for the baseline. This represents a 6.6\% improvement in failure detection. Moreover, its targeted generation strategies exposed a broader range of failure modes, particularly hallucinations triggered by unsupported queries, which were not detected by the baseline.
\end{minipage}
}
\end{center}

\subsection{RQ3 -- PDF Generation Strategies}

The manually selected Internet PDFs proved to be much stronger test cases than the GPT-5 Nano auto-generated PDFs. Tables \ref{tab:openai_mut_failures} and \ref{tab:ollama_mut_failures} show that the Manually-selected PDFs show substantially higher failure counts than the GPT 5-nano generated PDFs across all 24 combinations. In the case of OpenAI's models, \tool uncovered a total of 5,595 failures on Internet PDFs, significantly higher than the 3,285 AI-generated PDFs, that is, a 70.3\% increase. In the case of open-weight LLMs, \tool found 7,855 failures on Internet PDFs, higher than the 4,898 AI-generated PDFs. These results also translate to the baseline algorithm. 

A potential reason for these results might be that real-world Internet PDFs are significantly more challenging due to irregular structures, diverse writing styles, domain-specific jargon, noisy formatting, and subtle factual interdependencies that the GPT-5 Nano generator (even with carefully crafted prompts) does not fully replicate. This explains why they expose far more failures, they are simply stronger test material. The practical implication is clear: while the automated PDF generator is extremely convenient and enables fully reproducible, zero-effort test corpora, it currently underestimates risk compared to real documents. For production-grade RAG evaluation (especially in enterprise or high-stakes settings), teams should combine both strategies, i.e., use AI-generated PDFs for rapid iteration and scale, but always validate final results with a set of real sourced documents to catch the hardest failure modes. This also points to a clear future-work direction: enhance the PDF generator (e.g., by fine-tuning prompts on real PDFs, injecting controlled noise/variability, or adding multi-document synthesis) so that auto-generated corpora become as strong as real-world ones. RQ3 therefore shows that automated document generation is viable and scalable, but not yet a complete replacement for manually curated real documents.

We can therefore answer RQ3 as follows:

\begin{center}
\fbox{
\begin{minipage}{0.94\linewidth}
\textbf{Answer to RQ3.}
Real-world PDFs were substantially more effective at exposing failures than automatically generated PDFs across all 24 configurations. Overall, manually selected PDFs triggered 13,450 failures, compared with 8,183 for GPT-5 Nano-generated PDFs, corresponding to a 64.4\% increase. Thus, automatically generated documents provide a scalable and reproducible testing mechanism, but they do not yet fully reproduce the structural and semantic complexity of real-world documents.
\end{minipage}
}
\end{center}

\begin{table*}[h]
\centering
\small
\setlength{\tabcolsep}{6pt}
\renewcommand{\arraystretch}{1.2}
\caption{OpenAI MUT failures}

\resizebox{\textwidth}{!}{

\begin{tabular}{llcccccc}
\toprule
 &  & \multicolumn{3}{c}{\textbf{\tool}} & \multicolumn{3}{c}{\textbf{Baseline}} \\
\cmidrule(lr){3-5} \cmidrule(lr){6-8}
\textbf{Embedding} & \textbf{LLM} 
& AI & Internet & Total failures 
& AI & Internet & Total failures \\
\midrule

\multirow{4}{*}{\textbf{text-embedding-3-large}}
 & GPT 3.5-Turbo & 307   & 654   & 961    & 237   & 391   & 628 \\
 & GPT 4.1-Mini   & 198 & 310 & 508 & 127 & 241 & 368 \\
 & GPT 4o-Mini    & 246 & 446 & 692 & 182 & 330 & 512 \\
 & GPT 5-Nano     & 250 & 434 & 684  & 214 & 367 & 581 \\
\midrule

\multirow{4}{*}{\textbf{text-embedding-3-small}}
 & GPT 3.5-Turbo & 341   & 655   & 996    & 300   & 453   & 753 \\
 & GPT 4.1-Mini   & 241 & 354 & 595 & 187 & 306 & 493 \\
 & GPT 4o-Mini    & 304 & 451 & 755  & 232 & 390 & 622 \\
 & GPT 5-Nano     & 299 & 445 & 744 & 277 & 414 & 691 \\
\midrule

\multirow{4}{*}{\textbf{text-embedding-ada-002}}
 & GPT 3.5-Turbo & 321   & 574   & 895    & 290   & 441   & 731 \\
 & GPT 4.1-Mini   & 215 & 342 & 557 & 181 & 279 & 460 \\
 & GPT 4o-Mini    & 285 & 451 & 736 & 246 & 386 & 632 \\
 & GPT 5-Nano     & 278 & 479 & 757 & 262 & 399 & 661 \\
\bottomrule
\end{tabular}

}
\label{tab:openai_mut_failures}
\end{table*}

\begin{table*}[h]
\centering
\small
\setlength{\tabcolsep}{6pt}
\renewcommand{\arraystretch}{1.2}
\caption{Ollama MUT failures}

\resizebox{\textwidth}{!}{

\begin{tabular}{llcccccc}
\toprule
 &  & \multicolumn{3}{c}{\textbf{\tool}} & \multicolumn{3}{c}{\textbf{Baseline}} \\
\cmidrule(lr){3-5} \cmidrule(lr){6-8}
\textbf{Embedding} & \textbf{LLM} 
& AI & Internet & Total failures 
& AI & Internet & Total failures \\
\midrule

\multirow{4}{*}{\textbf{bge-large-335m-en-v1.5}}
 & Deepseek R1 & 265   & 488   & 753    & 288   & 381   & 669 \\
 & Llama 3.2   & 437 & 774 & 1211 & 758 & 982 & 1740 \\
 & Gemma 3    & 413 & 654 & 1067 & 364 & 596 & 960 \\
 & Qwen 3     & 378 & 605 & 983  & 371 & 541 & 912 \\
\midrule

\multirow{4}{*}{\textbf{intfloat-e5-base-v2}}
 & Deepseek R1 & 237   & 478   & 715    & 296   & 387   & 683 \\
 & Llama 3.2   & 614 & 904 & 1518 & 773 & 993 & 1766 \\
 & Gemma 3    & 381 & 615 & 996  & 360 & 650 & 1010 \\
 & Qwen 3     & 384 & 617 & 1001 & 365 & 571 & 936 \\
\midrule

\multirow{4}{*}{\textbf{nomic-embed-text-v1.5}}
 & Deepseek R1 & 301   & 493   & 794    & 306   & 393   & 699 \\
 & Llama 3.2   & 642 & 934 & 1576 & 762 & 1014 & 1776 \\
 & Gemma 3    & 435 & 670 & 1105 & 403 & 636 & 1039 \\
 & Qwen 3     & 411 & 623 & 1034 & 404 & 567 & 971 \\
\bottomrule
\end{tabular}

}
\label{tab:ollama_mut_failures}
\end{table*}


\subsection{RQ4 - Oracle Precision}

After applying the majority voting strategy, 302 tests were correctly identified as failures (True Positives), 74 were incorrectly flagged as failures when they had in fact passed (False Positives), and the remaining 6 produced ambiguous results where the LLMs could not reach a confident consensus. These 6 cases were reviewed manually and all classified as True Positives.

The Test Oracle achieved a precision of 80.6\%, meaning that roughly 4 out of every 5 tests it flags as failures are genuine. The remaining 19.4\% represent false positives, cases where the Oracle raised an alarm unnecessarily. While there is room for improvement, these results indicate that the Test Oracle performs reasonably well, correctly identifying the vast majority of true failures while keeping false alarms at a manageable level.

In conclusion, RQ4 can be answered as follows:

\begin{center}
\fbox{
\begin{minipage}{0.94\linewidth}
\textbf{Answer to RQ4.}
The test oracle achieved a precision of 80.6\%, correctly identifying approximately four out of every five reported failures. Specifically, 308 of the 382 inspected cases were confirmed as true failures, whereas 74 were false positives, resulting in a false discovery rate of 19.4\%. Although the oracle is sufficiently reliable for automated large-scale testing, its verdicts should be interpreted with caution when individual failure classifications are critical.
\end{minipage}
}
\end{center}

\section{Related Work}


Current RAG testing methods prioritize a component-level diagnostic strategy that isolates the retrieval and generation stages to pinpoint specific architectural failure modes. For retrieval testing, tools like RAGatouille \cite{AnswerDotAI} expose semantic mismatches or poor document ranking, and RanX \cite{10.1007/978-3-030-99739-7_30} measures low recall and low precision. To test the generation component, G-Eval \cite{liu-etal-2023-g} uses a ``Chain of Thought'' approach where a superior LLM acts like a human-like judge, effective to test fluency, tone and coherence. On the other hand, UpTrain \cite{Uptrain-Ai} heaviliy focuses on answer completeness and factual accuracy, by providing specific signals to tell if the generator provided a correct answer but missed key details that were present in the retrieved text.

This is predominantly operationalized through the RAG Triad, which evaluates context relevance (query-to-context), faithfulness or groundedness (context-to-answer) and answer relevance (query-to-answer), present in RAGAS~\cite{es-etal-2024-ragas}. Retrieval effectiveness can also be measured metrics like Recall@K \cite{patel2022recallksurrogatelosslarge}, Mean Reciprocal Rank (MRR) \cite{lewis2020rag}, and Normalized Discounted Cumulative Gain (NDCG) \cite{lewis2020rag}, while generation quality is assessed via ``LLM-as-a-judge'' frameworks that use advanced models to score semantic accuracy and detect hallucinations. Modern testing also incorporates automated red-teaming \cite{10.1145/3605764.3623985} and security scanning \cite{huang2024trustllmtrustworthinesslargelanguage} to proactively identify vulnerabilities such as prompt injections, biased outputs, and sensitive data leakage.

The frontier of RAG benchmarking has transitioned from simple question-answering to evaluating complex multi-step reasoning, multimodal integration, and corpus-level analysis. Established frameworks such as RGB~\cite{Chen_Lin_Han_Sun_2024} and CRUD-RAG~\cite{10.1145/3701228} evaluate fundamental capabilities like noise robustness and ``Create, Read, Update, Delete'' task taxonomy, while HaluEval~\cite{li-etal-2023-halueval} provides a comprehensive collection of human-annotated samples for detecting hallucinations. Recent research has introduced GlobalQA~\cite{luo2025globalretrievalaugmentedgeneration}, the first benchmark specifically designed for global reasoning tasks like corpus-wide sorting and counting, and mtRAG~\cite{10.1162/TACL.a.19}, which addresses the unique challenges of multi-turn conversational consistency. To ensure reliability in high-stakes domains, benchmarks like RAGguard~\cite{zeng2026worsezeroshotfactcheckingdataset} and RARE~\cite{zeng2025rareretrievalawarerobustnessevaluation} stress-test systems against misleading retrievals and document perturbations. Furthermore, emerging metrics such as UDCG (Utility and Distraction-aware Cumulative Gain)~\cite{trappolini-etal-2026-redefining} are being proposed to replace classical information retrieval metrics, as they demonstrate a significantly stronger correlation with end-to-end RAG accuracy.

Safety and robustness testing for Large Language Models (LLMs) has evolved into a rigorous discipline focused on ensuring models resist harmful instructions and remain stable under adversarial conditions. This is primarily operationalized through automated red teaming, where frameworks like HarmBench~\cite{mazeika2024harmbenchstandardizedevaluationframework} and JailbreakBench~\cite{NEURIPS2024_63092d79} systematically probe for vulnerabilities using hundreds of curated "misuse behaviors" and "jailbreak artifacts". Evaluation is further structured by specialized datasets and methodologies: ToxiGen~\cite{hartvigsen-etal-2022-toxigen} assesses nuanced, implicit toxicity; the Bias Benchmark for Question Answering (BBQ) ~\cite{parrish-etal-2022-bbq} measures social stereotypes across nine protected categories; Retromorphic Testing~\cite{yu2026retromorphictestinghierarchicalverification} introduces hierarchical verification to detect hallucinations specifically within Retrieval-Augmented Generation (RAG) workflows. Furthermore, advERSEM~\cite{dhole-etal-2025-adversem} tests for resilience against subtle linguistic manipulations like hedging. To align these technical tests with organizational standards, enterprises are increasingly adopting the NIST AI Risk Management Framework (AI RMF)~\cite{925786}, specifically utilizing its "Measure" function to validate seven key pillars of trustworthiness-including safety, fairness, and resilience-throughout the AI lifecycle.

More recently, Kim et al.~\cite{kim2026testing} introduced
\emph{Chunk Coverage} (CC), an oracle-independent test adequacy
criterion for the retrieval component of RAG systems. CC measures the proportion of corpus chunks retrieved at least once by a test suite and uses runtime retrieval feedback to guide the selection or generation of queries toward previously unexercised chunks. The approach therefore focuses on determining how thoroughly a test suite exercises the retrieval behavior induced by an embedding model and vector-store configuration, independently of the correctness of the generated answers. \tool addresses a complementary objective. Instead of isolating
the retriever, it automatically tests the complete retrieval-generation pipeline by generating retrievable documents, test queries and expected outputs, executing different combinations of LLMs and embedding models, and applying an automated test oracle to the resulting answers. Moreover, \tool generates different categories of failure-oriented tests, including queries targeting complex passages and negative queries
for which the system should reject unsupported requests. The coverage mechanisms also differ: CC measures which chunks are actually retrieved during test execution, whereas \tool's document-coverage strategy selects source regions that have not yet contributed to test generation. Thus, CC characterizes the adequacy of the exercised retrieval space, while our approach aims to expose failures arising from the interaction between retrieval and generation.

\tool implements a comprehensive "full-stack" evaluation lifecycle that closely mirrors the automated workflows of leading industrial frameworks like DeepEval~\cite{deepeval} and Maxim AI~\cite{maxim}. By automating the generation of synthetic test datasets from source documents (PDFs) and benchmarking various LLM and embedding configurations, our approach aligns with the industry's shift toward high-velocity RAGOps pipelines. However, while the current setup provides a robust end-to-end assessment based on the RAG Triad standards (faithfulness, relevance and groundedness), contemporary research solutions have specialized into deeper reasoning categories. For instance, where \tool tests general retrieval, GlobalQA~\cite{luo2025globalretrievalaugmentedgeneration} and CRUD-RAG~\cite{10.1145/3701228} introduce rigorous taxonomies for corpus-level operations like global sorting and multi-document summarization. Furthermore, the llm-as-a-judge mechanism could be further refined by adopting machine-oriented metrics like UDCG~\cite{trappolini-etal-2026-redefining}, which move beyond simple correctness to specifically quantify the "distraction" effect that irrelevant retrieved PDFs have on the model's final response, a critical safety dimension addressed by robustness benchmarks like RAGuard~\cite{zeng2026worsezeroshotfactcheckingdataset}.

\section{Conclusion and Future Work}

Testing Retrieval-Augmented Generation systems is challenging because their behavior depends on the interaction between document processing, retrieval, embeddings, and language generation. In this paper, we introduced \tool, an automated end-to-end approach that generates retrieval documents and test inputs, executes tests across different LLM embedding configurations, and evaluates the resulting answers using an LLM-based oracle.
We evaluated \tool on 24 LLM embedding configurations through 72,000 test executions. The approach detected 21,633 failures and outperformed the baseline in 20 of the 24 configurations, identifying 6.6\% more failures overall. The results also showed that real-world PDFs exposed substantially more failures than automatically generated documents, while the test oracle achieved a precision of 80.6\%. Overall, these findings indicate that systematic test generation can reveal retrieval inaccuracies, unsupported answers, incomplete context integration, and difficulties in processing complex passages that may remain unnoticed through conventional benchmark-based evaluation.

Future work will extend \tool to additional document formats and more complex query types, including multi-document, multi-hop, and adversarial test cases. We also plan to improve the test oracle, incorporate retrieval-specific metrics, and investigate more realistic document-generation strategies. Lastly, we will aim at incorporating CC~\cite{kim2026testing} as a novel adequacy metric within \tool

\section*{Replication Package}

\noindent The code of our approach can be found in \url{https://github.com/Anje13/GBL_RAGTester}.

\noindent The replication package can be found in \url{https://zenodo.org/records/21606024}.

\section*{Acknowledgments}

Jon Ayerdi, Miren Illarramendi and Aitor Arrieta are part of the Systems and Software Engineering research group of Mondragon Unibertsitatea (IT1919-26), supported by the Department of Education, Universities and Research of the Basque Country. 

\bibliographystyle{IEEEtran}
\bibliography{bibliografia}

\newpage
\section{Appendix}
\par\noindent\rule{\textwidth}{0.4pt}

\appendix
    
\section{PDF Generation System Prompt} 
\label{appendix:pdf_system}
\begin{lstlisting}
@staticmethod
def get_system_prompt(
    length: str = None,
    detail_level: str = None,
    tone: str = None,
    include_toc: bool = None,
    include_citations: bool = None
) -> str:

    length = length or PDFGeneratorPrompts.DEFAULT_LENGTH
    detail_level = detail_level or PDFGeneratorPrompts.DEFAULT_DETAIL_LEVEL
    tone = tone or PDFGeneratorPrompts.DEFAULT_TONE
    include_toc = include_toc if include_toc is not None else PDFGeneratorPrompts.INCLUDE_TOC
    include_citations = include_citations if include_citations is not None else PDFGeneratorPrompts.INCLUDE_CITATIONS
    
    system_prompt = f"""You are an expert technical writer and content creator specialized in generating comprehensive, well-structured documents.

    Your task is to create {detail_level} content suitable for conversion into a professional PDF document.

    DOCUMENT SPECIFICATIONS:
    - Target length: {length}
    - Writing style: {tone}
    - Target audience: {PDFGeneratorPrompts.DEFAULT_AUDIENCE}

    STRUCTURE REQUIREMENTS:
    {"- Include a detailed table of contents with section numbers" if include_toc else ""}
    {"- Include proper citations and references in academic format" if include_citations else ""}
    - Use clear headings and subheadings (H1, H2, H3)
    - Include numbered sections for easy reference
    - Break content into logical, digestible sections

    CONTENT REQUIREMENTS:
    - Provide in-depth explanations with technical accuracy
    - Include relevant examples and use cases where appropriate
    {"- Add tables, lists, and structured data where it aids understanding" if PDFGeneratorPrompts.INCLUDE_TABLES else ""}
    - Ensure factual accuracy and cite sources when making specific claims
    - Use clear, professional language
    - Maintain consistency in terminology throughout

    FORMAT GUIDELINES:
    - Use markdown formatting for structure
    - Clearly mark sections with ## for main sections, ### for subsections
    - Use bullet points and numbered lists for clarity
    - Include code blocks where relevant (use ```language syntax)
    - Use **bold** for emphasis and *italics* for technical terms

    QUALITY STANDARDS:
    - Ensure logical flow between sections
    - Provide sufficient depth on each topic
    - Avoid redundancy and repetition
    - Include transitional statements between major sections
    - Conclude with a comprehensive summary

    Generate content that is publication-ready and can be directly converted to PDF format."""

    return system_prompt
\end{lstlisting}

\section{PDF Generation User Prompt} 
\label{appendix:pdf_user}
\begin{lstlisting}
@staticmethod
def get_user_prompt(
    topic: str = None,
    specific_requirements: list = None,
    sections_to_include: list = None,
    sections_to_exclude: list = None,
    additional_context: str = None,
    doc_name : str = None
) -> str:

    user_prompt = f"""Please create a comprehensive document about{" a topic of your choice. You may choose from any category - science, arts, history, society, culture, nature, literature or any other area that inspires you. Try to be as diverse as you can and create very distinct documents." if topic is None else f": {topic}"}

    """
    
    # Add specific requirements if provided
    if specific_requirements:
        user_prompt += "SPECIFIC REQUIREMENTS:\n"
        for req in specific_requirements:
            user_prompt += f"- {req}\n"
        user_prompt += "\n"
    
    # Add required sections if provided
    if sections_to_include:
        user_prompt += "MUST INCLUDE THESE SECTIONS:\n"
        for section in sections_to_include:
            user_prompt += f"- {section}\n"
        user_prompt += "\n"
    
    # Add sections to exclude if provided
    if sections_to_exclude:
        user_prompt += "DO NOT INCLUDE:\n"
        for section in sections_to_exclude:
            user_prompt += f"- {section}\n"
        user_prompt += "\n"
    
    # Add additional context if provided
    if additional_context:
        user_prompt += f"ADDITIONAL CONTEXT:\n{additional_context}\n\n"
    
    user_prompt += f"""Please structure the document logically and ensure all content is accurate, well-researched, and suitable for a professional audience.
    Begin with a title page and table of contents, then proceed with the main content.{" Give the document a name in this EXACT format at the very end of the document. Make sure the document name does not include separate words, join them with '_' if necessary:" if doc_name is None else ""}
    {"DOCUMENT NAME: name" if doc_name is None else ""}"""

    return user_prompt
\end{lstlisting}

\section{Test Generation System Prompt} 
\label{appendix:testgen_system}
\begin{lstlisting}
@staticmethod
def get_system_prompt_types(
    question_type: str = None,
    num_questions_per_section: int = None
) -> str:

    question_specs = {
        "document maximizing": {
            "label": "DOC_MAX",
            "description": "Document Maximizing Questions",
            "rules": [
                "Cover different sections of the document",
                "Ensure broad coverage across all major topics",
                "Each question must target a distinct section or topic",
                "Questions should require retrieving information from specific document parts"
            ],
            "example": "What are the three main components described in Section 2?"
        },
        "negative rejection": {
            "label": "NEG_REJ",
            "description": "Negative Rejection Questions",
            "rules": [
                "Since no document content is provided for this case, generate question about any topic of your choice",
                "Ensure broad coverage across many topics",
                "Each question must target a distinct section or topic",
                "Questions should require retrieving information from specific document parts"
            ],
            "example": "What are the three main components described in Section 2?"
        },
        "NLP - flesch score": {
            "label": "LOW_FLESCH",
            "description": "Low Flesch Score Questions",
            "rules": [
                "Target subsections with the lowest Flesch readability scores",
                "Focus on complex, technical, or dense content",
                "Questions should test understanding of difficult passages",
                "Prioritize sections with specialized terminology or complex sentence structures"
            ],
            "example": "Explain the technical process described in the methodology section"
        },
        "cross section synthesis": {
            "label": "CROSS_SECTION",
            "description": "Cross Section Synthesis Questions",
            "rules": [
                "Require synthesizing information from two or more distinct sections or documents",
                "Each hop should contribute a necessary intermediate fact",
                "The answer must not be derivable from a single passage alone",
                "Favor questions that require chaining definitions, causes, or dependencies",
                "Avoid purely enumerative or extractive questions",
                "Ensure intermediate reasoning steps are implicit rather than explicitly stated together"
            ],
            "example": "How does the authentication mechanism described in the security section impact the rate-limiting behavior outlined in the API usage guidelines?"
        }
    }
    
    # Get the specific question type configuration
    if question_type not in question_specs:
        raise ValueError(f"Invalid question_type: {question_type}. Must be one of {list(question_specs.keys())}")
    
    spec = question_specs[question_type]
    
    system_prompt = f"""You are an expert test designer specialized in creating comprehensive test cases for evaluating RAG (Retrieval-Augmented Generation) systems.

    Your task is to analyze the given contents and generate high-quality question-answer pairs that will be used to test whether a RAG system can accurately retrieve and synthesize information.

    CRITICAL INSTRUCTION - READ CAREFULLY

    YOU MUST GENERATE EXACTLY {num_questions_per_section} QUESTIONS OF TYPE: {spec['label']} ({spec['description']})

    This is the ONLY question type you should generate in this session.
    DO NOT mix question types.
    DO NOT generate questions of any other category.
    ALL {num_questions_per_section} questions must have "question_type": "{spec['label']}"

    QUESTION TYPE SPECIFICATIONS: {spec['label']}

    {chr(10).join(f"  {i+1}. {rule}" for i, rule in enumerate(spec['rules']))}

    Example Question:
    {spec['example']}

    REQUIRED OUTPUT FORMAT

    For Each Test Case, You MUST Provide:

    1. **id**: Sequential number (1 to {num_questions_per_section})
    2. **question**: Clear, unambiguous question matching the {spec['label']} type
    3. **ground_truth_answer**: Accurate, concise answer based ONLY on document content
    4. **source_document**: Name/path of the source document, visible in section metadata as 'source'
    5. **source_location**: {"Specific section/page where answer is found. Visible in section metadata as 'page_number'" if question_type != "negative rejection" else "unknown"}
    6. **relevant_text_span**: Exact quote from document containing the answer
    7. **question_type**: MUST be exactly "{spec['label']}" for ALL questions
    8. **difficulty**: "easy", "medium", or "hard"
    9. **reasoning_required**: Brief description of reasoning needed
    10. **keywords**: Key terms from the answer for evaluation

    QUALITY STANDARDS

    Question Quality:
    - Avoid yes/no questions; prefer questions requiring specific information
    - Ensure questions are clear and have a single correct answer
    - Create questions that test different aspects of the document
    - Questions must align with the {spec['label']} category requirements

    Answer Quality:
    - Answers should be factual and verifiable in the document
    - Include specific details (numbers, dates, names) when relevant
    - Keep answers concise but complete (2-4 sentences typically)
    - Quote exact phrases from the document when appropriate
    {"- For negative rejection questions, answer MUST be: 'Insufficient information to answer this question'" if question_type == "negative rejection" else "Indicate if multiple pieces of information need to be synthesized"}

    OUTPUT FORMAT (JSON)

    Return a JSON array with EXACTLY {num_questions_per_section} objects in this format:

    [
    {{
        "id": 1,
        "question": Your question here,
        "ground_truth_answer": The answer is...,
        "source_document": folder/filename.pdf
        "source_location": Page X, text starting with... ,
        "relevant_text_span": Exact quote...,
        "question_type": {spec['label']},
        "difficulty": "easy", "medium" or "hard",
        "reasoning_required": Description of reasoning needed,
        "keywords": "keyword1, keyword2, keyword3"
    }},
    ... (continue for all {num_questions_per_section} questions)
    ]

    BEGIN GENERATION NOW - Remember: ONLY generate {spec['label']} type questions!
    """
    
    return system_prompt
\end{lstlisting}

\section{Test Generation User Prompt} 
\label{appendix:testgen_user}
\begin{lstlisting}
@staticmethod
def get_user_prompt(
    question_type: str = None,
    pdf_path: str = None,
    document_content: str = None,
    num_questions: int = 50,
    sorted_docs: list = None
) -> str:

    user_prompt = f"""Please analyze the following contents and generate test cases for evaluating a RAG system's ability to retrieve answer questions about this content.
    
    {(
    "CONTENT SPANS:"
    f"{document_content}"
    if (question_type == "document maximizing" or question_type == "cross section synthesis") and document_content else ""
    )}
    
    {(
    "SORTED SPANS:"
    f"{sorted_docs}"
    if question_type == "NLP - flesch score" and sorted_docs else ""
    )}
    """

    if num_questions:
        user_prompt += f"Generate {num_questions} test cases total.\n\n"

    user_prompt += """Ensure the test cases:
    1. Cover all major topics in the document
    2. Include a mix of difficulty levels
    3. Are answerable solely from the provided content
    4. Include exact source locations for verification

    Generate the test cases now in the specified JSON format."""

    return user_prompt
\end{lstlisting}

\section{Test Execution Prompt} 
\label{appendix:testexec_prompt}
\begin{lstlisting}
def create_prompt(self):
    prompt = PromptTemplate(
        template="""You are an assistant for question-answering tasks.
        Use the following documents to answer the question.
        If you don't know the answer, write 'Insufficient information to answer this question'.
        Make sure the answer you are giving is complete and no information is missing.
        Also include the name of the source document, the location of the answer in the original document and the relevant text span.
        Question: {question}
        Documents: {documents}

        Format your response as follows:
        {{
            "question": "",
            "answer": "",
            "source_document": "Name/path of the source document",
            "source_location": "Page X, text starting with...",
            "relevant_text_span": ""
        }}
        """,
        input_variables=["question", "documents"],
    )

    return prompt
\end{lstlisting}

\section{Test Oracle Prompt} 
\label{appendix:oracle_prompt}
\begin{lstlisting}
evaluation_prompt = PromptTemplate(
    template="""You are an expert judge evaluating a model's response against a ground truth standard.

    INPUTS:
    Query: {query}
    Question Type: {question_type}
    Model Answer: {answer} | Ground Truth Answer: {correct_answer}
    Model Source Document: {source_document} | Correct Source Document: {correct_document}
    Model Source Location : {source_location} | Correct Source Location: {correct_source}
    Model Text Span: {relevant_text_span} | Correct Text Span: {correct_span}

    Compare the model answer with the correct answer, as well as the source locations and selected relevant text spans.
    EVALUATION CRITERIA:
    1. Faithfulness (true - partially - false): Is the model answer grounded compared to the correct answer?
    2. Relevance (true - partially - false): Does the model answer directly address the query?
    3. Completeness (true - partially - false): Is the Model Under Test answer as complete as the correct one? If the Model Under Test answer is more complete than the correct answer, evaluate it positively.
    4. Clarity (true - partially - false): Is the generated answer clear and well-structured?
    5. Keywords match (true - partially - false): Do the keywords appear in the Model Under Test's answer? True if all keywords appear, partially if some appear and false if none appear.

    Write two lines summarizing the permorfance of the model under test. Compare the Correct Source Location with the Model Source Location, Correct Text Span with the Model Text Span and Correct Source Document with the Model Source Document.
    Also compare the Model Answer and the Ground Truth Answer, how the model answered differently compared to the correct answer, and mention if it the Model Answer is admittable or not. Justify each part.
    When calculating overall score, take true as 1, false as 0 and partially as 0.5.

    EVALUATION RULES:
        - If faithfulness is false, verdict should be FAIL.
        - If the model answer is empty or is not faithful to the ground truth answer, evaluation Faithfulness should be FAIL.
        - If question type is NEG_REJ and the Model Answer states there is insufficient information to answer the Question, verdict should be PASS, ignoring the source document, source location and relevant text span. In the evaluation mention that the model corretly admitted it was incapable to answer the question.
        - If question type is NEG_REG and the Model Answer states there is insufficient information to answer the Question, faithfulness, relevance, completeness and clarity should be true.
        - If question type is NEG_REJ and the model answers normally, compare it with the Ground Truth Answer and evaluate it as usual.

    Provide your evaluation in this EXACT JSON format:
    {json_template}

    Note: overall_score must be a number between 0.0 and 1.0

    Your evaluation:""",
    input_variables=["query", "question_type", "answer", "correct_answer", "source_document", "correct_document", "source_location", "correct_source", "relevant_text_span", "correct_span"]
)
\end{lstlisting}


\end{document}